\pdfoutput=1

\documentclass[11pt]{article}

\usepackage[final]{acl}

\usepackage{ragged2e}

\newcommand{\fooAlter}{\hspace{0pt}\textcolor{blue}{$\bullet$} \hspace{5pt}}

\usepackage{times}
\usepackage{latexsym}
\usepackage{paralist}

\usepackage[T1]{fontenc}

\usepackage[utf8]{inputenc}

\usepackage{microtype}

\usepackage{inconsolata}

\usepackage{graphicx}

\usepackage{booktabs}
\usepackage{csquotes}
\usepackage{enumitem}
\usepackage{colortbl}
\usepackage{multirow}
\usepackage{comment}
\usepackage{subcaption}

\title{Rater Cohesion and Quality from a Vicarious Perspective}

\author{
 \textbf{Deepak Pandita\textsuperscript{1}},
 \textbf{Tharindu Cyril Weerasooriya\textsuperscript{1}},
 \textbf{Sujan Dutta\textsuperscript{1}},\\
 \textbf{Sarah K. Luger\textsuperscript{2}},
 \textbf{Tharindu Ranasinghe\textsuperscript{3}},
 \textbf{Ashiqur R. KhudaBukhsh\textsuperscript{1}},\\
 \textbf{Marcos Zampieri\textsuperscript{4}},
 \textbf{Christopher M. Homan\textsuperscript{1}}
\\
\\
 \textsuperscript{1}Rochester Institute of Technology, USA,
 \textsuperscript{2}MLCommons, USA\\
 \textsuperscript{3}Lancaster University, UK,
 \textsuperscript{4}George Mason University, USA
\\
\small{
\{deepak, cyril, khudabukhsh\}@mail.rit.edu, sarah@mlcommons.org,}\\
\small{t.ranasinghe@lancaster.ac.uk, \{sd2516, cmhvcs\}@rit.edu}
}

\begin{document}

\maketitle

\begin{abstract}

\textcolor{red}{This paper discusses and contains offensive content.} 
Human feedback is essential for building human-centered AI
systems across domains where disagreement is prevalent, such as AI safety, content moderation, or sentiment analysis. Many disagreements, particularly in politically charged settings, arise because raters have opposing values or beliefs.
\emph{Vicarious annotation} is a method for breaking down disagreement by asking raters how they think others would annotate the data. In this paper, we explore the use of vicarious annotation with analytical methods for moderating rater disagreement.
We employ rater-cohesion metrics to study the potential influence of political affiliations and demographic backgrounds on raters' perceptions of offense. Additionally, we utilize \emph{CrowdTruth}'s rater quality metrics, which consider the demographics of the raters, to score the raters and their annotations. We study how the rater-quality metrics influence the in-group and cross-group rater cohesion across the personal and vicarious levels.
\end{abstract}

\section{Introduction}

A crucial part of many AI systems is the humans who provide feedback for learning or evaluation \citep{vaughan_making_2018}.
As AI systems grow more powerful, aligning models with human values becomes even more critical. Recent work in reinforcement learning with human feedback (RLHF) 
\citep{pmlr-v70-macglashan17a,lin_review_2020,ouyang_training_2022,casper2023open}
highlights the gains in model performance from aligning them to human values. The research into RLHF also notes the technical challenges associated with doing so. 

A major challenge to eliciting human feedback is that raters frequently disagree with each other \citep{uma_learning_2021}. Annotating political discourse is particularly challenging because disagreements are tied to human raters' values \citep{jost2009political}, making disagreement in political domains more explicit than in other annotation tasks \citep{yano-etal-2010-shedding,lukin-etal-2017-argument,sap-etal-2022-annotators,weerasooriya-etal-2023-vicarious}.

\citet{weerasooriya-etal-2023-vicarious} introduced the concept of \emph{vicarious offense}, where human raters are asked to annotate data according to their own opinions, and also \emph{vicariously}, e.g., on behalf of specific groups to which they \emph{do not} belong. Such \emph{vicarious annotations} can reveal whether a group can be trusted to represent the opinions of other groups. If the group can be trusted, then we can recruit fewer raters from the other groups and still have enough annotations to represent the population from which the raters are drawn. If the group cannot be trusted, we need to find another group that can be trusted; otherwise, the only way to obtain a representative set of annotations is to recruit from all groups. 
 
This paper explores group coherence in vicarious annotation tasks investigating the following research questions.
\vspace{-2mm}

\paragraph{RQ1} \textit{Are some groups more cohesive than others when disclosing their own perceptions of offense?} 
\paragraph{RQ2} \textit{How much variance is there among the cohesion levels observed when different groups predict vicarious offense for other groups?}
\paragraph{RQ3}  \textit{What is the impact of removing raters deemed low-quality by CrowdTruth on group cohesion?}

We address questions using the metrics introduced in the GRASP framework \citep{prabhakaran-etal-2024-grasp} for understanding rater cohesion and CrowdTruth \citep{Dumitrache_SAD_CROWDBIAS_HCOMP2018}, two approaches for measuring the impacts of rater disagreement. Rater cohesion metrics measure the extent to which rater disagreement is based on group membership. CrowdTruth teases disagreement apart due to differences of opinion from poor rater quality.

The major takeaways are that, of the political groups, Independents are the most cohesive, both with themselves and with others. Democrats are the least cohesive with others. Republicans are the least internally cohesive.

\section{Related Work}

Prior work has highlighted the prevalence of disagreement in aggregated labels for subjective NLP tasks such as toxic language detection \citep{binns2017like, park-etal-2018-reducing, sap-etal-2019-risk, davidson-etal-2019-racial, al-kuwatly-etal-2020-identifying}. Disagreement is often due to rater identity (race, gender, age, education, and first language) and their beliefs (political leaning) \citep{sap-etal-2019-risk, al-kuwatly-etal-2020-identifying, larimore-etal-2021-reconsidering, sap-etal-2022-annotators, goyal2022your, pei-jurgens-2023-annotator, weerasooriya-etal-2023-vicarious, homan-etal-2024-intersectionality, prabhakaran-etal-2024-grasp}. Studies have also highlighted the impact of rater bias on NLP datasets \citep{geva-etal-2019-modeling}. To uncover and analyze these differences, previous work has relied on regression models and training classifiers using demographic information and comparing their predictions \citep{binns2017like, davidson-etal-2019-racial, al-kuwatly-etal-2020-identifying, larimore-etal-2021-reconsidering, goyal2022your}.

Recent work has advocated the use of non-aggregated (rater-level) labels  \citep{basile-etal-2021-need, prabhakaran-etal-2021-releasing, plank-2022-problem, Cabitza_Campagner_Basile_2023} to enable an extensive treatment of this variation. To this end, \citet{homan-etal-2024-intersectionality} used Bayesian multilevel models to discover intersectional effects between rater demographics and their ratings. \citet{prabhakaran-etal-2024-grasp} proposed GRASP (Group Associations in Perspectives), a framework to analyze (dis)agreement among rater subgroups. CrowdTruth \citep{Dumitrache_SAD_CROWDBIAS_HCOMP2018} is another framework that benefits from rater-level labels for evaluating the quality of a dataset through three dimensions: individual raters, input data items, and the overall dataset.

Our approach has one similarity to \emph{Bayesian truth serum}, (BTS), where ``impersonally informative'' questions garner more honest answers 
\citep{Prelec2004} when these answers are gathered in groups.
In practice, this means using pairs of questions where one question asks for the individual's opinion and the second asks them to estimate the group distribution for this question.
``BTS relies on the Bayesian assumption that people maintain a mental model of the world that is biased by their personal experiences, which leads to a belief that personally held opinions are disproportionately present amongst peers'' \citep{Frank2017}.
By asking raters from one political group to consider what they believe another political group thinks, we replicate the first part of BTS methodology by getting distanced, and thus more honest perceptions of how the original rater perceives a topic. Another difference is BTS works less optimally for judging subjective social posts because it requires experts to agree on a single truth, not multiple valid truths. 
BTS (and its variants) are used in various crowd-sourcing projects to effectively gather more honest self-reported data on non-subjective topics \citep{witkowski2012robust, faltings2014incentives, Frank2017}. 

\section{Methods}

\subsection{Vicarious annotation}

Given a dataset of machine learning training or test items $\mathcal{X}$, a rater pool $\mathcal{Z}$, a subgroup $Z$ of $\mathcal{Z}$, and a question $q$ with response domain $\mathcal{D}$ that is asked of each item a \emph{vicarious annotation} of $\mathcal{X}$ with respect to $(q, \mathcal{D}, Z, \mathcal{Z})$ is a matrix $\mathbf{Y}_Z$ having one row for each item in $\mathcal{X}$ and one column for each rater $z \not\in Z$, and entries in $\mathcal{D}$, where the entries are responses to the question -- \emph{How would a rater in $Z$ annotate $q$?} 

\subsection{Group Cohesion Metrics}
We use the GRASP framework \cite{prabhakaran-etal-2024-grasp} and the associated metrics to compare in-group cohesion and cross-group divergences. This framework utilizes permutation tests along with proposed metrics to measure the variability of judgments by diverse rater subgroups. The metrics are either \emph{in-group} and \emph{cross-group}. All metrics are designed so that larger values mean more cohesion.

\subsubsection{In-group Metrics}
In-group metrics (indicated by $\cap$'s) measure the cohesion among raters within a group. Each metric captures a slightly different aspect of cohesion, and together, they form a robust signal of group cohesion.

\noindent\fooAlter\textbf{IRR:} Inter-rater reliability (IRR) is used to measure agreement among multiple raters in a way that controls for class imbalance in the distribution of ratings over all items. Specifically, we use Krippendorff's alpha \citep{krippendorff2004reliability}, a
metric that generalizes several IRR metrics by accepting an arbitrary number of raters, different levels of measurement, handling missing data, and adjusting to small sample sizes.

\noindent\fooAlter\textbf{Negentropy:} Negentropy \citep{brillouin1953negentropy}, unlike IRR, does not control for class imbalance, but it does account for the entire distribution of rater responses for each item. It is computed by subtracting for each item the entropy over responses from the maximum value entropy can take. Then, we compute the mean over all the items.

\noindent\fooAlter\textbf{Plurality size:} Plurality size is the fraction of raters that belong to the majority vote. Traditionally, gold standard data is based on the plurality choice for each item. However, it only measures cohesion among the most popular choice for each item; it ignores the rest of the responses, in contrast to the previous two metrics.

\subsubsection{Cross-group Metrics}
Cross-group metrics (indicated by $\otimes$'s) measure the cohesion between the raters belonging to different groups. Each of these metrics roughly corresponds to an in-group metric.

\noindent\fooAlter\textbf{XRR:} Cross-replication reliability \cite{wong-etal-2021-cross} is similar to IRR but is defined for raters from different groups. 

\noindent\fooAlter\textbf{Cross negentropy:} Cross negentropy is similar to negentropy but is computed over two distributions. 

\noindent\fooAlter\textbf{Voting agreement:} Voting agreement is similar to plurality size and is computed by taking the most popular response for each item for each group and then calculating Krippendorff's alpha between the two groups.

\subsubsection{Group Association Index}
GAI combines in-group and cross-group cohesion into a single score. We define GAI as the ratio of IRR to XRR. Thus, values higher than 1 would indicate higher in-group cohesion while values smaller than 1 indicate higher cross-group cohesion. A value of 1 indicates low or no group association.

\subsection{CrowdTruth}
CrowdTruth (CT) \citep{Dumitrache_SAD_CROWDBIAS_HCOMP2018} is a framework that connects three key dimensions: individual raters, input data items, and annotations. These dimensions are interconnected in the CT algorithm to prevent situations where disagreement from low-quality raters can reduce the overall data quality or ambiguous data leads to poor rater performance. The quality of the raters is influenced by the quality of the data items they have annotated and the quality of the annotations in the dataset. In this study, we calculate the CT for the entire dataset with the rater demographics and focus on the individual rater quality scores. The relevant score for the research is the worker quality score (WQS), which measures the overall agreement of one rater with other raters. We use a publicly available implementation\footnote{https://github.com/CrowdTruth/CrowdTruth-core} of CT.\footnote{Our code is available at \url{https://github.com/Homan-Lab/rater_cohesion_public/}}

\section{Data}

\subsection{Datasets}
We apply GRASP and CrowdTruth frameworks to $\mathcal{D}_\textit{voiced}$ \citep{weerasooriya-etal-2023-vicarious}, a collection of 2,338 comments on YouTube videos from the official channels of three leading US cable news networks (CNN, FOX, and MSNBC) labeled by diverse raters for offensiveness. Collected over eight years spanning from 2014 to 2022, the comments, therefore, cover a variety of topics~\cite{khudabukhsh2021we}. Refer to Tables \ref{tab:dvo_stats}--\ref{tab:dvo_rater_polxgender} for more information about the dataset. We consider political leaning and gender as dimensions to compare agreement among different subgroups. This dataset is unique as it also contains labels for vicarious offenses, where raters are asked to predict offenses for others who do not share their political beliefs. We also inspect these vicarious labels of offense and compare the cohesion across subgroups.

In addition, we also apply the frameworks to the toxicity ratings dataset \citep{kumar2021designing} of 107,620 comments from Twitter, Reddit, and 4chan labeled for toxicity by 17,280 raters. The dataset contains demographic information about the raters and their political leaning but does not contain vicarious annotations. We sample 250 batches of 20 comments, resulting in a total of 5,000 items ($\mathcal{D}_\textit{toxicity}$ from hereafter). The sampling process ensures that each batch is annotated by the same five raters, consistent with the original dataset. We then remove raters who report multiple political affiliations. Refer to Tables \ref{tab:dvo_stats} and \ref{tab:dtr_rater_polxgender} for more information about the dataset.

\subsection{Examples}
We now present three examples from $\mathcal{D}_\textit{voiced}$ to highlight our key findings. See Appendix \ref{metric_study} for additional examples and further discussion.

\vspace{-1mm}
\noindent\rule{0.48\textwidth}{1pt}
\raggedright{\noindent
\textbf{EXP1} (MSNBC): \textit{Send trump and his deranged softy sons to Iran!!!}}
\noindent\rule{0.48\textwidth}{1pt}

\justifying
The majority vote for \textbf{EXP1} across groups is offensive. Plurality scores: Democrats: 0.83, Independents: 0.87, Republicans: 0.83, Women: 0.91, Men: 0.75, Men\textsuperscript{IND}: 0.75, Women\textsuperscript{IND}: 1.0, and Men\textsuperscript{REP}: 0.5. \textbf{Findings}: Even though the comment is targeted towards sons of the former US President Donald Trump, there is a portion of men as a whole and even republican leaning men finding it not offensive.

\noindent\rule{0.48\textwidth}{1pt}
\raggedright\noindent
\textbf{EXP2} (Fox): \textit{Pretty soon they will start counting abortions at covid deaths. Fricking Dems}
\noindent\rule{0.48\textwidth}{1pt}

\justifying
The majority vote for \textbf{EXP2} across groups is offensive. Plurality scores: Democrats: 0.66, Independents: 1, Republicans: 0.78, Women: 0.76, Men: 1.0, and Women\textsuperscript{DEM}: 0.50. Other gender\textsuperscript{PP} had a score of 1. \textbf{Findings}: In contrast to \textbf{EXP1}, \textbf{EXP2} contains a targeted attack on the Democrats, yet only 0.66 of the Democrats agreed that it is offensive, and women were less cohesive in contrast to men.

\noindent\rule{0.48\textwidth}{1pt}
\raggedright\noindent
\textbf{EXP3} (CNN): \textit{fuck abortion dude is literally murder bruh why do they support so fucking bad killing a child like what is wrong with those ppl smh!! dude it takes Man and a woman to create babies}
\noindent\rule{0.48\textwidth}{1pt}

\justifying
The majority vote for \textbf{EXP3} across groups is offensive. Plurality scores: Democrats: 0.66, Independents: 1, Republicans: 0.77, Women: 0.76, Men: 1, Women\textsuperscript{DEM}: 0.50, and Women\textsuperscript{REP}: 0.60. Other gender\textsuperscript{PP} had a score of 1. \textbf{Findings}: men are cohesive in their overall opinion, however, women are not as strongly cohesive as men.

\section{Results}

Tables \ref{tab:tb_metrics}, \ref{tab:tb_metrics_tr_250}, \ref{tab:tb_vic_metrics}, \ref{tab:tb_ct_metrics}, \ref{tab:tb_ct_metrics_tr_250}, and \ref{tab:tb_vic_ct_metrics} show results for in-group (indicated by $\cap$'s) and cross-group (indicated by $\otimes$'s) cohesion for the personal and vicarious offense.
Significant results (see Appendix \ref{sig_testing} for details) are indicated in bold at the $p=0.05$ significance level, $\downarrow$ indicates the result is less than expected under the null hypothesis and $\uparrow$ indicates the result is greater than expected.

\subsection{Group Cohesion for Personal Offense}

\begin{table*}[h]
\centering
\small
\begin{tabular}{r|ccccccc}
& & & & Cross $\otimes$ & Plurality $\cap$ & Voting $\otimes$ & \\
Group & IRR $\cap$ & XRR $\otimes$ & Negentropy $\cap$ & Negentropy & Size & Agreement & GAI \\
\hline
Dem & $\uparrow$0.176 & $\downarrow$0.148 & $\downarrow$0.323 & \textbf{$\downarrow$0.257} & $\downarrow$0.815 & \textbf{$\downarrow$0.321} & $\uparrow$1.193 \\
Rep & $\downarrow$0.139 & $\downarrow$0.154 & $\downarrow$0.326 & $\downarrow$0.282 & $\downarrow$0.824 & $\downarrow$0.394 & $\downarrow$0.902 \\
Ind & $\uparrow$0.208 & $\uparrow$0.178 & \textbf{$\uparrow$0.433} & $\uparrow$0.310 & \textbf{$\uparrow$0.872} & \textbf{$\uparrow$0.488} & $\uparrow$1.171 \\
\hline
Men & $\uparrow$0.178 & \textbf{$\downarrow$0.149} & $\uparrow$0.338 & $\downarrow$0.301 & $\uparrow$0.834 & $\downarrow$0.414 & $\uparrow$1.196 \\
Women & $\downarrow$0.156 & \textbf{$\downarrow$0.150} & $\downarrow$0.308 & $\uparrow$0.317 & $\downarrow$0.817 & $\downarrow$0.410 & $\uparrow$1.044 \\
\hline
Dem, Men & $\uparrow$0.204 & $\uparrow$0.177 & $\uparrow$0.484 & $\uparrow$0.310 & $\uparrow$0.884 & $\uparrow$0.335 & $\uparrow$1.152 \\
Dem, Women & $\uparrow$0.167 & $\downarrow$0.161 & $\downarrow$0.391 & \textbf{$\downarrow$0.179} & \textbf{$\downarrow$0.826} & \textbf{$\downarrow$0.085} & $\uparrow$1.042 \\
Rep, Men & $\downarrow$0.108 & $\downarrow$0.150 & $\downarrow$0.421 & $\uparrow$0.280 & $\downarrow$0.853 & $\downarrow$0.308 & $\downarrow$0.725 \\
Rep, Women & $\uparrow$0.170 & \textbf{$\downarrow$0.118} & $\downarrow$0.410 & $\downarrow$0.206 & $\downarrow$0.851 & $\downarrow$0.215 & \textbf{$\uparrow$1.434} \\
Ind, Men & $\uparrow$0.203 & $\uparrow$0.184 & $\downarrow$0.457 & $\downarrow$0.249 & $\downarrow$0.868 & $\uparrow$0.277 & $\uparrow$1.103 \\
Ind, Women & $\downarrow$0.154 & $\downarrow$0.149 & \textbf{$\uparrow$0.567} & \textbf{$\uparrow$0.375} & \textbf{$\uparrow$0.925} & $\uparrow$0.377 & $\uparrow$1.029 \\
\end{tabular}
\caption{Results of in-group and cross-group cohesion metrics on $\mathcal{D}_\textit{voiced}$. $\cap$ stands for in-group metric and $\otimes$ stands for cross-group metric. Significant results are indicated in bold at the $p=0.05$ significance level, $\downarrow$ indicates the result is less than expected under the null hypothesis, and $\uparrow$ indicates the result is greater than expected.}
\label{tab:tb_metrics}
\end{table*}

\paragraph{Results on $\mathcal{D}_\textit{voiced}$:} Table \ref{tab:tb_metrics} shows results on $\mathcal{D}_\textit{voiced}$ for personal-level offense by political leaning and gender. For each group, we report in-group metrics for the group and cross-group metrics between the group and all raters not in the group.
For political leaning, only Independents show uniformly higher in-group and cross-group agreement than other groups. Democrats have significantly lower cross-group agreement but mixed in-group agreement results. They also have the highest GAI score. Republicans have no significant results, but all agreement metrics and GAI are lower than median random scores.
 
For gender, men and women have significantly lower XRR scores than random groups. Intersectional groups of political leaning and gender show some noteworthy differences from the single-variable groups and some extreme values, particularly among women. Both Republican and Democratic women have lower-than-expected in- and cross-group scores. Particularly notable is that the voting agreement for Democratic women is  0.085, compared to 0.321 for Democrats and 0.415 for all women, and is by far the lowest among all groups, single or intersectional. Republican women, by contrast, have the highest GAI score (1.434) among all groups. Independent women generally have higher than expected agreement scores, with the highest Negentropy, Cross Negentropy, and Plurality size of all groups by a substantial margin.

\begin{table*}[h]
\centering
\small
\begin{tabular}{r|ccccccc}
& & & & Cross $\otimes$ & Plurality $\cap$ & Voting $\otimes$ & \\
Group & IRR $\cap$ & XRR $\otimes$ & Negentropy $\cap$ & Negentropy & Size & Agreement & GAI \\
\hline
Dem & $\uparrow$0.283 & $\downarrow$0.258 & $\uparrow$0.545 & $\downarrow$0.494 & $\uparrow$0.905 & $\downarrow$0.287 & $\uparrow$1.097 \\
Rep & \textbf{$\downarrow$0.185} & \textbf{$\downarrow$0.237} & $\uparrow$0.596 & $\downarrow$0.450 & $\uparrow$0.933 & \textbf{$\downarrow$0.233} & $\downarrow$0.783 \\
Ind & $\uparrow$0.292 & $\downarrow$0.266 & $\uparrow$0.610 & $\downarrow$0.444 & $\uparrow$0.942 & $\uparrow$0.306 & $\uparrow$1.097 \\
\hline
Men & $\downarrow$0.235 & $\downarrow$0.259 & $\uparrow$0.527 & $\downarrow$0.506 & $\uparrow$0.897 & $\downarrow$0.289 & $\downarrow$0.905 \\
Women & $\uparrow$0.283 & $\downarrow$0.251 & $\uparrow$0.502 & $\uparrow$0.538 & $\uparrow$0.879 & $\downarrow$0.277 & $\uparrow$1.128 \\
\hline
Dem, Men & \textbf{$\downarrow$0.157} & $\downarrow$0.247 & $\uparrow$0.635 & $\downarrow$0.422 & $\uparrow$0.959 & $\downarrow$0.276 & $\downarrow$0.635 \\
Dem, Women & $\uparrow$0.303 & \textbf{$\uparrow$0.299} & $\downarrow$0.602 & $\downarrow$0.430 & $\downarrow$0.937 & $\uparrow$0.334 & $\uparrow$1.013 \\
Rep, Men & $\downarrow$0.155 & \textbf{$\downarrow$0.221} & $\downarrow$0.639 & \textbf{$\downarrow$0.379} & $\downarrow$0.961 & \textbf{$\downarrow$0.223} & $\downarrow$0.703 \\
Rep, Women & $\uparrow$0.287 & $\downarrow$0.240 & \textbf{$\uparrow$0.664} & $\uparrow$0.424 & \textbf{$\uparrow$0.980} & \textbf{$\downarrow$0.251} & $\uparrow$1.199 \\
Ind, Men & $\uparrow$0.395 & $\downarrow$0.262 & $\uparrow$0.654 & \textbf{$\uparrow$0.446} & $\uparrow$0.972 & $\downarrow$0.266 & $\uparrow$1.506 \\
Ind, Women & $\downarrow$0.220 & $\uparrow$0.282 & $\uparrow$0.648 & $\uparrow$0.419 & $\uparrow$0.967 & $\uparrow$0.356 & $\downarrow$0.781 \\
\end{tabular}
\caption{Results of in-group and cross-group cohesion metrics on $\mathcal{D}_\textit{toxicity}$. $\cap$ stands for in-group metric and $\otimes$ stands for cross-group metric. Significant results are indicated in bold at the $p=0.05$ significance level, $\downarrow$ indicates the result is less than expected under the null hypothesis and $\uparrow$ indicates the result is greater than expected.}
\label{tab:tb_metrics_tr_250}
\end{table*}

\paragraph{Results on $\mathcal{D}_\textit{toxicity}$:} Table \ref{tab:tb_metrics_tr_250} shows results on $\mathcal{D}_\textit{toxicity}$ for personal-level offense by political leaning and gender. For political leaning, Republicans show significantly lower IRR, XRR, and voting agreement scores than random groups. Democrats show higher-than-expected in-group scores and lower-than-expected cross-group scores. Independents have higher-than-expected in-group scores.

Intersectional groups of political leaning and gender reveal some noteworthy differences. Republican women have higher-than-expected in-group scores with the highest Negentropy (0.664) and Plurality size (0.980) among all groups. In contrast, Republican men have lower-than-expected in- and cross-group scores. Particularly notable is that Republican men have the lowest scores among all groups for all cross-group metrics. They also have the lowest IRR (0.155) among all groups. Democrat men have lower-than-expected cross-group scores with the lowest GAI (0.635). Independent men have higher-than-expected in-group scores with the highest IRR (0.395) and GAI (1.506) among all groups, while Independent women have higher-than-expected cross-group scores with the highest voting agreement (0.356) among all groups.

Compared to Table \ref{tab:tb_metrics}, Democrats, Republicans, and men remain consistent with lower-than-expected cross-group scores. Independents remain consistent with higher-than-expected in-group scores and Republican men with lower-than-expected in-group scores. Women flip from their lower-than-expected in-group score to a higher-than-expected score. Democrat men flip from their higher-than-expected cross-group score to a lower-than-expected score.

\subsection{Group Cohesion for Vicarious Offense}

\begin{table*}[h]
\centering
\small
\begin{tabular}{r|ccccccc}
& & & & Cross $\otimes$ & Plurality $\cap$ & Voting $\otimes$ & \\
Group & IRR $\cap$ & XRR $\otimes$ & Negentropy $\cap$ & Negentropy & Size & Agreement & GAI \\
\hline
Rep $\rightarrow$ Dem (v Dem) & $\downarrow$0.164 & $\downarrow$0.140 & \textbf{$\downarrow$0.386} & $\downarrow$0.330 & \textbf{$\downarrow$0.855} & $\uparrow$0.330 & $\downarrow$1.169 \\
Ind $\rightarrow$ Dem (v Dem) & $\uparrow$0.220 & $\uparrow$0.183 & $\downarrow$0.460 & $\downarrow$0.349 & $\downarrow$0.886 & $\uparrow$0.353 & $\downarrow$1.201 \\
\hline
Dem $\rightarrow$ Rep (v Rep) & $\uparrow$0.172 & $\downarrow$0.127 & \textbf{$\downarrow$0.300} & \textbf{$\downarrow$0.259} & \textbf{$\downarrow$0.797} & $\uparrow$0.247 & $\uparrow$1.350 \\
Ind $\rightarrow$ Rep (v Rep) & $\uparrow$0.188 & $\uparrow$0.163 & $\uparrow$0.425 & $\uparrow$0.343 & $\uparrow$0.863 & \textbf{$\uparrow$0.354} & $\downarrow$1.153 \\
\hline
Dem $\rightarrow$ Ind (v Ind) & $\uparrow$0.143 & $\downarrow$0.145 & \textbf{$\downarrow$0.328} & \textbf{$\uparrow$0.416} & \textbf{$\downarrow$0.815} & $\downarrow$0.268 & $\uparrow$0.982 \\
Rep $\rightarrow$ Ind (v Ind) & $\uparrow$0.141 & $\uparrow$0.171 & $\downarrow$0.347 & \textbf{$\uparrow$0.421} & $\downarrow$0.832 & $\uparrow$0.323 & $\downarrow$0.824 \\
\end{tabular}
\caption{Results of vicarious alignment on $\mathcal{D}_\textit{voiced}$. $\cap$ stands for in-group metric and $\otimes$ stands for cross-group metric. Significant results are indicated in bold at the $p=0.05$ significance level, $\downarrow$ indicates the result is less than expected under the null hypothesis and $\uparrow$ indicates the result is greater than expected.}
\label{tab:tb_vic_metrics}
\end{table*}

First, looking at in-group cohesion of vicarious predictions (e.g., Republican $\rightarrow$ Democrat) shown in Table \ref{tab:tb_vic_metrics} versus self-ratings from the predicting group (e.g., Republican) shown in Table \ref{tab:tb_metrics}, Republicans are more cohesive by all in-group metrics when predicting vicariously, rather than for themselves. This is also true for Democrats, but only when predicting for Independents. When predicting for Republicans in-group cohesion numbers are mixed. Independents show very similar results to Democrats.

Next, looking at in-group cohesion of vicarious predictions (e.g., Republican $\rightarrow$ Democrat, Table \ref{tab:tb_vic_metrics}) to self-ratings from the target group (e.g., Democrats, \ref{tab:tb_metrics}), Independents have greater in-group cohesion when predicting for either Republicans or Democrats than for themselves. Both Republicans and Democrats show inconclusive results.

Finally, Table \ref{tab:tb_vic_metrics} compares self-ratings from the target group (e.g., Democrats) to vicarious predictions (e.g., Republican $\rightarrow$ Democrat). Independents have a higher cohesion with Democrats when predicting vicarious offense for Democrats than Republicans predicting vicarious offense for them by all cross-group metrics. Independents also have higher cohesion with Republicans when predicting vicariously for Republicans than Democrats predicting vicariously for Republicans by all cross-group metrics. Particularly noteworthy is that the voting agreement for Independents predicting vicariously for Republicans is significantly higher (0.354) as compared to Democrats predicting vicariously for Republicans which is the lowest among all groups (0.247). Republicans have a higher cohesion with Independents when predicting vicariously for them than Democrats predicting vicariously for Independents by all cross-group metrics.

\subsection{CrowdTruth Evaluation}
As introduced earlier, CrowdTruth's triangle of disagreement is dependent on the raters, data item/unit, and annotations. We focus on the worker quality score (WQS, ranging from a minimum of 0 to 1) for this study. The WQS measures the overall agreement of one rater over other raters and favors raters that agree with others. 

\begin{figure}[htb]
    \centering
    \begin{subfigure}{\linewidth}
    \includegraphics[width=\linewidth]{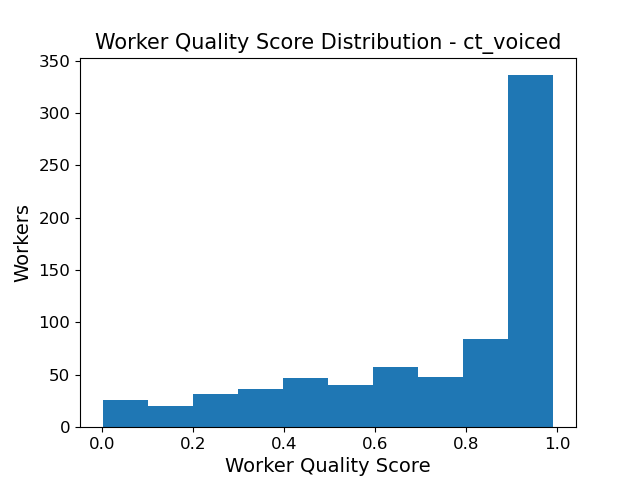}
    \end{subfigure}
    \begin{subfigure}{\linewidth}
    \includegraphics[width=\linewidth]{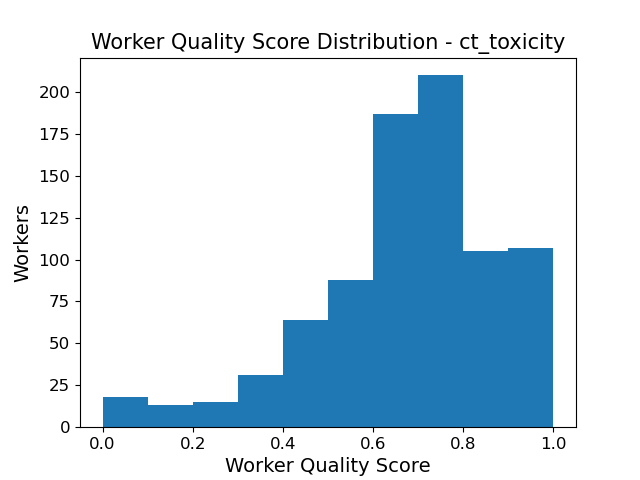}
    \end{subfigure}
    \caption{Distribution of CrowdTruth worker quality score (WQS) for each rater in the datasets. We use the WQS to filter out lower-rated raters from the datasets.}
    \label{fig:ct_wqs}
\end{figure}

Figure~\ref{fig:ct_wqs} shows the distribution of WQS from CrowdTruth. We filter out raters with a WQS below 0.1 and re-run our cohesion metrics on the remaining dataset. We identified WQS 0.1 as a cut-off based on the distribution of the scores. 

Table \ref{tab:crowdtruth-removed} shows the data impacted after filtering out raters deemed low-quality using CrowdTruth. Nearly identical numbers of Republican-, Democrat-, and Independent-leaning raters were removed (8--9 each) from $\mathcal{D}_\textit{voiced}$. More Republican-leaning raters (12) were removed from $\mathcal{D}_\textit{toxicity}$ than Democrat- and Independent-leaning raters (3 and 2, respectively).

\begin{table}
\centering
  \begin{tabular}{r|r|r}
  & $\mathcal{D}_\textit{voiced}$ & $\mathcal{D}_\textit{toxicity}$\\
  \hline
   Annotations & 2250 & 360\\
   Data items & 1405 & 360\\
   Democrats & 8 & 3\\
   Republicans & 9 & 12\\
   Independents & 8 & 2
\end{tabular}
\caption{Data impacted after CrowdTruth filtering.}
\label{tab:crowdtruth-removed}
\end{table}

\subsubsection{Results of Group Cohesion for Personal Offense after CrowdTruth Filtering}

\begin{table*}[h]
\centering
\small
\begin{tabular}{r|ccccccc}
& & & & Cross $\otimes$ & Plurality $\cap$ & Voting $\otimes$ & \\
Group & IRR $\cap$ & XRR $\otimes$ & Negentropy $\cap$ & Negentropy & Size & Agreement & GAI \\
\hline
Dem & $\uparrow$0.238 & $\downarrow$0.197 & $\downarrow$0.403 & $\downarrow$0.349 \cellcolor{orange} & $\downarrow$0.855 & \textbf{$\downarrow$0.367} \cellcolor{green} & \textbf{$\uparrow$1.203} \cellcolor{cyan} \\
Rep & $\downarrow$0.167 & $\downarrow$0.193 & $\downarrow$0.376 & $\uparrow$0.381 & $\downarrow$0.851 & $\downarrow$0.473 & $\downarrow$0.864 \\
Ind & $\uparrow$0.251 & $\uparrow$0.215 & \textbf{$\uparrow$0.487} \cellcolor{green} & $\uparrow$0.383 & \textbf{$\uparrow$0.898} \cellcolor{green} & $\uparrow$0.537 \cellcolor{orange} & $\uparrow$1.165 \\
\midrule
Men & $\uparrow$0.213 & \textbf{$\downarrow$0.187} \cellcolor{green} & $\uparrow$0.387 & $\downarrow$0.384 & $\uparrow$0.861 & $\downarrow$0.493 & $\uparrow$1.141 \\
Women & $\downarrow$0.202 & \textbf{$\downarrow$0.187} \cellcolor{green} & $\downarrow$0.379 & $\uparrow$0.384 & $\downarrow$0.854 & $\downarrow$0.482 & $\uparrow$1.085 \\
\hline
Dem, Men & $\uparrow$0.204 & $\uparrow$0.205 & $\downarrow$0.484 & $\uparrow$0.359 & $\downarrow$0.884 & $\downarrow$0.340 & $\uparrow$0.993 \\
Dem, Women & \textbf{$\uparrow$0.305} \cellcolor{cyan} & $\uparrow$0.222 & $\downarrow$0.507 & $\downarrow$0.302 \cellcolor{orange} & $\downarrow$0.892 \cellcolor{orange} & \textbf{$\downarrow$0.206} \cellcolor{green} & \textbf{$\uparrow$1.373} \cellcolor{cyan} \\
Rep, Men & $\downarrow$0.148 & $\downarrow$0.197 & $\uparrow$0.481 & $\uparrow$0.371 & $\uparrow$0.885 & $\uparrow$0.371 & $\downarrow$0.750 \\
Rep, Women & $\downarrow$0.175 & \textbf{$\downarrow$0.154} \cellcolor{green} & \textbf{$\downarrow$0.433} \cellcolor{cyan} & $\downarrow$0.299 & \textbf{$\downarrow$0.864} \cellcolor{cyan} & $\downarrow$0.272 & $\uparrow$1.142 \cellcolor{orange} \\
Ind, Men & $\uparrow$0.284 & $\uparrow$0.241 & $\uparrow$0.537 & $\downarrow$0.348 & $\uparrow$0.910 & $\uparrow$0.349 & $\uparrow$1.182 \\
Ind, Women & \textbf{$\downarrow$0.110} \cellcolor{cyan} & $\downarrow$0.174 & $\uparrow$0.572 \cellcolor{orange} & \textbf{$\uparrow$0.423} \cellcolor{green} & $\uparrow$0.930 \cellcolor{orange} & $\uparrow$0.393 & \textbf{$\downarrow$0.631} \cellcolor{cyan} \\
\hline
$\Delta$ & 0.047 & 0.041 & 0.053 & 0.083 & 0.029 & 0.060 & 0.130 \\
\end{tabular}
\caption{Results of in-group and cross-group cohesion metrics on $\mathcal{D}_\textit{voiced}$ after CrowdTruth filtering. $\cap$ stands for in-group metric and $\otimes$ stands for cross-group metric. Significant results are indicated in bold at the $p=0.05$ significance level, $\downarrow$ indicates the result is less than expected under the null hypothesis, and $\uparrow$ indicates the result is greater than expected. \colorbox{orange}{Orange} indicates the result is significant before applying CT, \colorbox{cyan}{Cyan} indicates the result is significant after applying CT, and \colorbox{green}{Green} indicates the result is significant before and after applying CT. $\Delta$ is the mean absolute difference of metric scores before and after applying CT.}
\label{tab:tb_ct_metrics}
\end{table*}

\begin{table*}[h]
\centering
\small
\begin{tabular}{r|ccccccc}
& & & & Cross $\otimes$ & Plurality $\cap$ & Voting $\otimes$ & \\
Group & IRR $\cap$ & XRR $\otimes$ & Negentropy $\cap$ & Negentropy & Size & Agreement & GAI \\
\hline
Dem & $\downarrow$0.293 & $\downarrow$0.289 & $\uparrow$0.548 & $\uparrow$0.518 & $\uparrow$0.907 & $\downarrow$0.306 & $\uparrow$1.013 \\
Rep & $\uparrow$0.307 \cellcolor{orange} & \textbf{$\downarrow$0.275} \cellcolor{green} & \textbf{$\uparrow$0.618} \cellcolor{cyan} & \textbf{$\downarrow$0.448} \cellcolor{cyan} & $\uparrow$0.948 & \textbf{$\downarrow$0.272} \cellcolor{green} & $\uparrow$1.119 \\
Ind & $\downarrow$0.294 & $\downarrow$0.292 & $\uparrow$0.611 & $\downarrow$0.460 & $\uparrow$0.943 & $\downarrow$0.325 & $\uparrow$1.009 \\
\hline
Men & $\downarrow$0.291 & $\downarrow$0.291 & $\uparrow$0.544 & $\downarrow$0.515 & $\uparrow$0.906 & $\downarrow$0.309 & $\uparrow$1.003 \\
Women & $\downarrow$0.297 & $\downarrow$0.285 & $\downarrow$0.508 & $\uparrow$0.552 & $\downarrow$0.883 & $\downarrow$0.303 & $\uparrow$1.042 \\
\hline
Dem, Men & \textbf{$\downarrow$0.157} \cellcolor{green} & $\downarrow$0.284 & $\uparrow$0.634 & $\uparrow$0.447 & $\downarrow$0.958 & $\downarrow$0.298 & \textbf{$\downarrow$0.552} \cellcolor{cyan} \\
Dem, Women & $\uparrow$0.303 & $\uparrow$0.318 \cellcolor{orange} & $\downarrow$0.601 & $\downarrow$0.449 & $\downarrow$0.937 & $\uparrow$0.344 & $\downarrow$0.953 \\
Rep, Men & $\uparrow$0.372 & $\downarrow$0.274 \cellcolor{orange} & $\uparrow$0.661 & \textbf{$\downarrow$0.389} \cellcolor{green} & $\uparrow$0.977 & $\downarrow$0.287 \cellcolor{orange} & $\uparrow$1.361 \\
Rep, Women & $\uparrow$0.357 & $\downarrow$0.277 & \textbf{$\uparrow$0.668} \cellcolor{green} & $\uparrow$0.439 & \textbf{$\uparrow$0.983} \cellcolor{green} & $\downarrow$0.282 \cellcolor{orange} & $\uparrow$1.287 \\
Ind, Men & $\uparrow$0.395 & $\downarrow$0.297 & $\uparrow$0.653 & \textbf{$\uparrow$0.458} \cellcolor{green} & $\uparrow$0.971 & $\downarrow$0.305 & $\uparrow$1.330 \\
Ind, Women & $\downarrow$0.220 & $\downarrow$0.294 & $\uparrow$0.648 & $\uparrow$0.442 & $\downarrow$0.967 & $\uparrow$0.354 & $\downarrow$0.747 \\
\hline
$\Delta$ & 0.044662 & 0.032115 & 0.007043 & 0.015255 & 0.004648 & 0.026464 & 0.162877 \\
\end{tabular}
\caption{Results of in-group and cross-group cohesion metrics on $\mathcal{D}_\textit{toxicity}$ after CrowdTruth filtering. $\cap$ stands for in-group metric and $\otimes$ stands for cross-group metric. Significant results are indicated in bold at the $p=0.05$ significance level, $\downarrow$ indicates the result is less than expected under the null hypothesis, and $\uparrow$ indicates the result is greater than expected. \colorbox{orange}{Orange} indicates the result is significant before applying CT, \colorbox{cyan}{Cyan} indicates the result is significant after applying CT, and \colorbox{green}{Green} indicates the result is significant before and after applying CT. $\Delta$ is the mean absolute difference of metric scores before and after applying CT.}
\label{tab:tb_ct_metrics_tr_250}
\end{table*}

\paragraph{Results on $\mathcal{D}_\textit{voiced}$:} Table \ref{tab:tb_ct_metrics} shows the results on $\mathcal{D}_\textit{voiced}$ for in-group and cross-group cohesion metrics for the dimensions of political leaning and gender after CrowdTruth filtering. Compared to Table \ref{tab:tb_metrics}, overall, nearly all in- and cross-group metrics increase after applying CrowdTruth. The most noteworthy exception is that Independent women have lower IRR and GAI scores. The number of significant results added and removed between tables after CrowdTruth filtering are all the same (seven). This number is approximately the same as the expected false positive rate at $p=0.05$, taking both Tables \ref{tab:tb_metrics} and \ref{tab:tb_ct_metrics} into account.
This result illustrates why we should not read the $p$-values here as a measure of statistical significance \emph{per se}, but rather as a relative measure of the likelihood the effects are due to a true difference in the underlying population from (a) random group(s) of the same size(s).

\paragraph{Results on $\mathcal{D}_\textit{voiced}$:} Table \ref{tab:tb_ct_metrics_tr_250} shows the results on $\mathcal{D}_\textit{toxicity}$ for in-group and cross-group cohesion metrics for the dimensions of political leaning and gender after CrowdTruth filtering. Compared to Table \ref{tab:tb_metrics_tr_250}, nearly all in- and cross-group metrics increase after applying CrowdTruth. Particularly notable is that Republicans and Republican men have higher-than-expected in-group scores and lower-than-expected cross-group scores.

\subsubsection{Results of Group Cohesion for Vicarious Offense after CrowdTruth Filtering}

Table \ref{tab:tb_vic_ct_metrics} shows the results for in-group and cross-group cohesion metrics for vicarious predictions after CrowdTruth filtering. Compared to Table \ref{tab:tb_vic_metrics}, overall, nearly all in- and cross-group metrics increase after applying CrowdTruth except for some minor variation in GAI.

\begin{table*}[h]
\centering
\small
\begin{tabular}{r|ccccccc}
& & & & Cross $\otimes$ & Plurality $\cap$ & Voting $\otimes$ & \\
Group & IRR $\cap$ & XRR $\otimes$ & Negentropy $\cap$ & Negentropy & Size & Agreement & GAI \\
\hline
Rep $\rightarrow$ Dem (v Dem) & $\downarrow$0.181 & $\downarrow$0.176 & \textbf{$\downarrow$0.419} \cellcolor{green} & $\downarrow$0.411 & \textbf{$\downarrow$0.871} \cellcolor{green} & $\downarrow$0.331 & $\downarrow$1.027 \\
Ind $\rightarrow$ Dem (v Dem) & $\uparrow$0.252 & \textbf{$\uparrow$0.231} \cellcolor{cyan} & $\downarrow$0.502 & $\downarrow$0.423 & $\uparrow$0.906 & \textbf{$\uparrow$0.418} \cellcolor{cyan} & $\downarrow$1.091 \\
\hline
Dem $\rightarrow$ Rep (v Rep) & $\uparrow$0.230 & $\downarrow$0.166 & \textbf{$\downarrow$0.376} \cellcolor{green} & \textbf{$\downarrow$0.346} \cellcolor{green} & \textbf{$\downarrow$0.840} \cellcolor{green} & $\uparrow$0.283 & $\uparrow$1.389 \\
Ind $\rightarrow$ Rep (v Rep) & $\uparrow$0.215 & $\uparrow$0.191 & $\uparrow$0.470 & $\downarrow$0.402 & $\uparrow$0.887 & \textbf{$\uparrow$0.393} \cellcolor{green} & $\downarrow$1.123 \\
\hline
Dem $\rightarrow$ Ind (v Ind) & $\uparrow$0.203 & $\uparrow$0.200 & $\downarrow$0.413 \cellcolor{orange} & \textbf{$\uparrow$0.487} \cellcolor{green} & $\downarrow$0.860 \cellcolor{orange} & $\uparrow$0.353 & $\uparrow$1.016 \\
Rep $\rightarrow$ Ind (v Ind) & $\downarrow$0.164 & $\uparrow$0.200 & \textbf{$\downarrow$0.393} \cellcolor{cyan} & \textbf{$\uparrow$0.486} \cellcolor{green} & $\downarrow$0.857 & $\uparrow$0.372 & $\downarrow$0.821 \\
\hline
$\Delta$ & 0.036 & 0.039 & 0.055 & 0.073 & 0.029 & 0.046 & 0.060 \\
\end{tabular}
\caption{Results of vicarious alignment on $\mathcal{D}_\textit{voiced}$ after CrowdTruth filtering. $\cap$ stands for in-group metric and $\otimes$ stands for cross-group metric. Significant results are indicated in bold at the $p=0.05$ significance level, $\downarrow$ indicates the result is less than expected under the null hypothesis, and $\uparrow$ indicates the result is greater than expected. \colorbox{orange}{Orange} indicates the result is significant before applying CT, \colorbox{cyan}{Cyan} indicates the result is significant after applying CT, and \colorbox{green}{Green} indicates the result is significant before and after applying CT. $\Delta$ is the mean absolute difference of metric scores before and after applying CT.}
\label{tab:tb_vic_ct_metrics}
\end{table*}

\section{Discussion}
Regarding RQ1, the major takeaways are that, of the political groups, Independents are the most cohesive, both with themselves and with others. Democrats are the least cohesive with others. Republicans are the least internally cohesive.

Because Independents split their votes between Democrats and Republicans in most elections, we were not surprised by their relatively high level of cohesion with other groups. However, their internal cohesion was not as commonsensical. Perhaps it was due to the Democrats and Republicans containing both extremist and more moderate members, who tend to agree on inoffensive content but whose extreme members are more readily ``triggered'' by moderately offensive content. And perhaps Independents contain fewer extreme members.
Regarding intersections between gender and political leaning, considering women raters mostly amplifies the results seen by political leaning.  
Independent women have the highest cohesion among themselves as well as with other groups. Democrat women have lower cohesion among themselves and with other groups. 
Our results suggest that women are driving disagreement among Democrats and agreement among Independents.

Regarding RQ2, Independents have higher cohesion with Democrats and Republicans while predicting vicariously for them.
Republicans have a higher cohesion with Independents while predicting vicarious offense. Democrats, again, appear to be the most isolated, this time in terms of their ability to predict what other groups find offensive.

For RQ3, it is not surprising that using CrowdTruth to remove raters leads to higher cohesion scores; the CrowdTruth quality metrics depend on within-group agreement levels. One notable exception is Independent women. 
In terms of p-value, XRR is the most stable metric. CrowdTruth filtering seems to particularly benefit IRR and XRR, and hurt the Negentropy metrics the most.

\subsection{Implications for Data Collection}
Since Independents are the most cohesive, one might conclude that with a limited budget, it would make sense to have slightly more Independents than other raters, because of their high cross-group cohesiveness. However, one must weigh this utilitarian conclusion against the cold reality that offensive content is sometimes directed at marginal groups by other marginal groups in such a way as to be unnoticeable by most people. And so, particularly in settings where such \emph{dog whistling} behavior is likely, this is likely the wrong action to take~\cite{mendelsohn-etal-2023-dogwhistles}. 

The relatively low level of cohesion between Democrats and other raters suggests that some minimum amount of the budget should be allocated to Democrats, because the other raters do not represent their beliefs. However, the relatively low level of cohesion between Democrats predicting vicariously for others suggests that they should not get too much budget. The relatively low levels of in-group cohesion for Women Democrats suggest they should have a larger substantial portion of the Democratic budget than men because there is more variance in their responses. 

A critical open question remains: how many raters do we need in each group for each item to have confidence that our annotations fairly represent the target audience? This is an important question that we believe has yet to get the attention it truly deserves.

\section{Conclusion}
Our investigation into the dynamics of rater cohesion in politically charged content moderation settings, through the lens of self and \textit{vicarious annotation}, gender, and political affiliations, reveals valuable insights into the challenges of building inclusive and human-centered AI systems. Our findings reveal notable disparities in cohesion levels, highlighting the influence of gender and political affiliation. For instance, Independent women and Democrat women show significantly different patterns of cohesion both within their groups and with other groups. We also note that Independents show higher vicarious cohesion with other groups. This finding opens up a strategic avenue for more efficient rater recruitment, implying that Independents can effectively approximate the viewpoints of Democrats and Republicans. Consequently, \textit{vicarious annotation} emerges as a valuable tool for optimizing rater recruitment, ensuring diverse representation under resource constraints.

\section{Limitations}
While our study computes subgroup cohesion metrics along two critical demographic dimensions (gender and political leaning), the findings may not be generalizable to other demographics such as education level, cultural background, and economic status. Future studies should employ the proposed framework to investigate the level of cohesion among raters belonging to other important demographic subgroups. Another limitation of this work is the simplification of political ideologies into three groups: Democrats, Republicans, and Independents. This, however, may not capture the full spectrum of political beliefs and identities. For instance, a rater can be socially Republican but fiscally Liberal. A more granular analysis that considers the multidimensional nature of political ideologies could reveal more intricate patterns of cohesion.

CrowdTruth is inherently an algorithm designed to dissolve disagreements. By filtering out lower-scored raters, we can remove disagreements, resulting in a more agreeable dataset.

\section*{Ethics Statement}
The datasets utilized in this study consist of a human-annotated compilation of publicly accessible YouTube, Twitter, Reddit, and 4chan comments, as introduced by \citet{kumar2021designing} and \citet{weerasooriya-etal-2023-vicarious}. These datasets do not reveal any identifiable information about the raters. The authors of the original datasets claimed they consulted their institutional review board to ensure the safety of raters during the data collection. We are aware that in some previous studies, raters had raised concerns about the impact of mental trauma when annotating for safety for ChatGPT \citep{hao_cleaning_2023} and social networks \citep{wexler_lawsuits_2023}. However, no such concerns were reported by the authors of these datasets.

\section*{Acknowledgements}

We thank the anonymous reviewers for their valuable feedback and suggestions on our work.

Marcos Zampieri is supported by the Virginia Commonwealth Cyber Initiative (CCI) award number N-4Q24-009.

\bibliography{anthology,custom,cyril_refs}

\appendix

\section{Appendix}
\label{sec:appendix}

\subsection{Dataset Statistics}
Tables \ref{tab:dvo_stats}--\ref{tab:dtr_rater_polxgender} show the annotation statistics and the raters in political leaning X gender intersectional groups for $\mathcal{D}_\textit{voiced}$ and $\mathcal{D}_\textit{toxicity}$.

\begin{table}[h]
\centering
\small
\begin{tabular}{c|cc}
& $\mathcal{D}_\textit{voiced}$ & $\mathcal{D}_\textit{toxicity}$\\
\hline
Items & 2338 & 5000\\
Raters & 726 & 803\\
Raters per item & \multirow{2}{*}{(5, 19, 60)} & \multirow{2}{*}{(1, 3, 5)}\\
(min, median, max) &&\\
Annotations & 45725 & 16380\\
\end{tabular}
\caption{Dataset annotation statistics}
\label{tab:dvo_stats}
\end{table}

\begin{table}[h]
\centering
\small
\begin{tabular}{r|rrr|r}
& Dem & Rep & Ind & Total \\
\hline
Men & 126 & 146 & 113 & 385\\
Women & 118 & 115 & 103 & 336\\
NA & 3 & 1 & 1 & 5 \\
\hline
Total & 247 & 262 & 217 & 726
\end{tabular}
\caption{$\mathcal{D}_\textit{voiced}$ raters in political leaning X gender intersectional groups}
\label{tab:dvo_rater_polxgender}
\end{table}

\begin{table}[h]
\centering
\small
\begin{tabular}{r|rrr|r}
& Dem & Rep & Ind & Total \\
\hline
Men & 147 & 114 & 108 & 369\\
Women & 201 & 116 & 111 & 428\\
NA & 4 & 2 & 0 & 6 \\
\hline
Total & 352 & 232 & 219 & 803
\end{tabular}
\caption{$\mathcal{D}_\textit{toxicity}$ raters in political leaning X gender intersectional groups}
\label{tab:dtr_rater_polxgender}
\end{table}

\subsection{Significance Testing}
\label{sig_testing}
Following \citet{prabhakaran-etal-2024-grasp}, we utilize \textit{null hypothesis significance tests (NHSTs)} to test the significance of our results. For any cohesion metric our null hypothesis $\bf H_{null}$ is that the value of the in-group or cross-group metric for any subgroup is independent of the political leaning and demographics of the raters.

We conduct permutation tests by randomly shuffling the political leaning and demographics of the raters, measuring the shuffled test statistic, and counting the number of times the shuffled statistic exceeds (lags) the observed value. For us, p-value is the fraction of times we observe a test statistic that is more extreme than the observed value. We use 1000 trials for our experiments.

Given the large number of tests we conduct, the value of NHST for us is not in the tests \emph{per se}, and so we do not correct for the false discovery rate. Rather, we follow \citet{10.1214/11-STS356} and others, who advocate using p-values in exploratory settings as a concise way of measuring the \emph{relative} significance of some results versus others.

\subsection{Metric-Guided Heuristic Study}
\label{metric_study}
We sampled examples from $\mathcal{D}_\textit{voiced}$ based on the cohesion metrics to understand the strengths and limitations of this work. We discussed EXP1 through EXP3, highlighting how each targeted demographic group shows the least cohesion. This theme exists for cases such as EXP5 and EXP6. 

EXP5 is an example where the comment is not attributing "Republic" to the Republicans but as a call to action to galvanize fellow Americans to vote for gun control and take action.  
Only 0.71 of the Republicans agree with this perspective, and women are even less cohesive. 

EXP7 is a case where the comment is offensive to the Democrat-leaning voters. However, they are the least cohesive out of the three political leanings. This further supports the observation of impacted/called-out demographic groups being less cohesive. 







\vspace{-1mm}
\noindent\rule{0.48\textwidth}{1pt}
\textbf{EXP4} (MSNBC): \textit{I love me some liberal tears! Let's Go Brandon!!!}
\noindent\rule{0.48\textwidth}{1pt}
\raggedright\noindent

\vspace{-1mm}
\justifying
The majority vote for \textbf{EXP4} across groups is offensive. Plurality scores: Democrats: 0.50, Independents: 0.85, Republicans: 0.57, Women: 0.50, Men: 0.87, and Women\textsuperscript{REP}: 0.75. Other gender\textsuperscript{PP} had a score of 1. 

\noindent\rule{0.48\textwidth}{1pt}
\raggedright\noindent
\textbf{EXP5} (MSNBC): \textit{Absolutely useless posts! Here are the facts: URL These corrupt politicians and lobbyist need to go! Get out and vote for safer gun laws in November for Senate Seat and in 2020!! WE the Republic are in charge -not these clowns! Remember, it could be your loved one next!!}
\noindent\rule{0.48\textwidth}{1pt}
\raggedright\noindent

\vspace{-1mm}
\justifying
The majority vote for \textbf{EXP5} across groups except for Democrat-leaning women is offensive. Plurality scores: Democrats: 0.57, Independents: 1, Republicans: 0.71, Women: 0.57, Men: 0.87, Women\textsuperscript{DEM}: 0.60, and Men\textsuperscript{REP}: 0.83. Other gender\textsuperscript{PP} had a score of 1.

\noindent\rule{0.48\textwidth}{1pt}
\textbf{EXP6} (FOX): \textit{If a person kills a pregnant mother, do they not get charged for a double homicide?}
\noindent\rule{0.48\textwidth}{1pt}

\vspace{-1mm}
\justifying
The majority vote for \textbf{EXP6} across groups is offensive. Plurality scores: Democrats: 0.60, Independents: 1, Republicans: 0.80, Women: 0.50, Men: 1, Women\textsuperscript{DEM}: 0.66, and Women\textsuperscript{REP}: 0.50. Other gender\textsuperscript{PP} had a score of 1.

\noindent\rule{0.48\textwidth}{1pt}
\raggedright\noindent
\textbf{EXP7} (MSNBC): \textit{heres a great idea why dont the rich and stupid old biden get rid of the guns on there body guards first before they try telling Americans to give up there weapons lets go brandan stupid dems beed to be thrown in prison they are traiters to this country}
\noindent\rule{0.48\textwidth}{1pt}

\vspace{-1mm}
\justifying
The majority vote for \textbf{EXP7} across groups is offensive. Plurality scores; Democrats: 0.66, Independents: 1, Republicans: 0.80, Women: 0.50, Men: 1, and Women\textsuperscript{DEM}: 0.80. Other gender\textsuperscript{PP} had a score of 1.


\subsection{Median permutation test scores}

\begin{table*}[h]
\centering
\small
\begin{tabular}{
    r|ccccccc
}
& & & & Cross $\otimes$ & Plurality $\cap$ & Voting $\otimes$ & \\
Group & {IRR $\cap$} & {XRR $\otimes$} & {Negentropy $\cap$} & {Negentropy} & {Size} & {Agreement} & {GAI} \\
\hline
Dem & 0.159 & 0.162 & 0.354 & 0.287 & 0.833 & 0.414 & 0.984 \\
Rep & 0.160 & 0.162 & 0.347 & 0.289 & 0.832 & 0.424 & 0.983 \\
Ind & 0.160 & 0.162 & 0.372 & 0.284 & 0.839 & 0.393 & 0.989 \\
\hline
Man & 0.159 & 0.163 & 0.316 & 0.315 & 0.824 & 0.456 & 0.980 \\
Woman & 0.161 & 0.163 & 0.325 & 0.303 & 0.826 & 0.456 & 0.991 \\
\hline
Dem, Man & 0.155 & 0.163 & 0.445 & 0.271 & 0.864 & 0.286 & 0.961 \\
Dem, Woman & 0.158 & 0.162 & 0.458 & 0.271 & 0.870 & 0.274 & 0.983 \\
Rep, Man & 0.157 & 0.163 & 0.422 & 0.272 & 0.856 & 0.310 & 0.969 \\
Rep, Woman & 0.155 & 0.162 & 0.461 & 0.271 & 0.871 & 0.271 & 0.967 \\
Ind, Man & 0.158 & 0.162 & 0.465 & 0.272 & 0.874 & 0.271 & 0.988 \\
Ind, Woman & 0.160 & 0.162 & 0.481 & 0.271 & 0.880 & 0.259 & 0.994 \\
\end{tabular}
\caption{Median permutation scores for in-group and cross-group cohesion on $\mathcal{D}_\textit{voiced}$}
\label{tab:tb_median_metrics}
\end{table*}

\begin{table*}
\centering
\small
\begin{tabular}{
    r|ccccccc
}
& & & & Cross $\otimes$ & Plurality $\cap$ & Voting $\otimes$ & \\
Group & {IRR $\cap$} & {XRR $\otimes$} & {Negentropy $\cap$} & {Negentropy} & {Size} & {Agreement} & {GAI} \\
\hline
Dem & 0.270 & 0.271 & 0.532 & 0.501 & 0.897 & 0.298 & 0.997 \\
Rep & 0.271 & 0.271 & 0.590 & 0.454 & 0.931 & 0.301 & 1.001 \\
Ind & 0.269 & 0.272 & 0.596 & 0.449 & 0.935 & 0.304 & 0.991 \\
\hline
Men & 0.270 & 0.270 & 0.526 & 0.507 & 0.894 & 0.299 & 1.000 \\
Women & 0.270 & 0.270 & 0.498 & 0.532 & 0.878 & 0.299 & 0.996 \\
\hline
Dem, Men & 0.270 & 0.272 & 0.629 & 0.426 & 0.956 & 0.304 & 0.997 \\
Dem, Women & 0.270 & 0.271 & 0.603 & 0.444 & 0.939 & 0.303 & 0.997 \\
Rep, Men & 0.275 & 0.272 & 0.644 & 0.415 & 0.966 & 0.307 & 1.001 \\
Rep, Women & 0.270 & 0.272 & 0.644 & 0.415 & 0.966 & 0.305 & 1.004 \\
Ind, Men & 0.271 & 0.270 & 0.647 & 0.412 & 0.968 & 0.304 & 1.003 \\
Ind, Women & 0.272 & 0.271 & 0.646 & 0.413 & 0.967 & 0.306 & 1.008 \\
\end{tabular}
\caption{Median permutation scores for in-group and cross-group cohesion on $\mathcal{D}_\textit{toxicity}$}
\label{tab:tb_median_metrics_tr}
\end{table*}

\begin{table*}
\centering
\small
\begin{tabular}{
    r|ccccccc
}
& & & & Cross $\otimes$ & Plurality $\cap$ & Voting $\otimes$ & \\
Group & {IRR $\cap$} & {XRR $\otimes$} & {Negentropy $\cap$} & {Negentropy} & {Size} & {Agreement} & {GAI} \\
\hline
Rep $\rightarrow$ Dem (v Dem) & 0.184 & 0.150 & 0.458 & 0.370 & 0.881 & 0.296 & 1.201 \\
Ind $\rightarrow$ Dem (v Dem) & 0.184 & 0.153 & 0.480 & 0.369 & 0.888 & 0.283 & 1.203 \\
\hline
Dem $\rightarrow$ Rep (v Rep) & 0.168 & 0.136 & 0.402 & 0.331 & 0.846 & 0.227 & 1.223 \\
Ind $\rightarrow$ Rep (v Rep) & 0.166 & 0.137 & 0.422 & 0.329 & 0.854 & 0.212 & 1.222 \\
\hline
Dem $\rightarrow$ Ind (v Ind) & 0.137 & 0.147 & 0.392 & 0.357 & 0.844 & 0.268 & 0.922 \\
Rep $\rightarrow$ Ind (v Ind) & 0.137 & 0.148 & 0.386 & 0.358 & 0.842 & 0.271 & 0.921 \\
\end{tabular}
\caption{Median permutation scores for vicarious alignment on $\mathcal{D}_\textit{voiced}$}
\label{tab:tb_vic_median_metrics}
\end{table*}

\begin{table*}
\centering
\small
\begin{tabular}{
    r|ccccccc
}
& & & & Cross $\otimes$ & Plurality $\cap$ & Voting $\otimes$ & \\
Group & {IRR $\cap$} & {XRR $\otimes$} & {Negentropy $\cap$} & {Negentropy} & {Size} & {Agreement} & {GAI} \\
\hline
Dem & 0.202 & 0.204 & 0.414 & 0.367 & 0.864 & 0.471 & 0.988 \\
Rep & 0.201 & 0.203 & 0.409 & 0.371 & 0.864 & 0.484 & 0.985 \\
Ind & 0.202 & 0.203 & 0.427 & 0.363 & 0.868 & 0.449 & 0.990 \\
\hline
Man & 0.201 & 0.203 & 0.377 & 0.394 & 0.856 & 0.512 & 0.988 \\
Woman & 0.203 & 0.203 & 0.389 & 0.383 & 0.859 & 0.512 & 1.001 \\
\hline
Dem, Man & 0.199 & 0.202 & 0.488 & 0.350 & 0.889 & 0.342 & 0.980 \\
Dem, Woman & 0.202 & 0.202 & 0.507 & 0.349 & 0.897 & 0.328 & 0.996 \\
Rep, Man & 0.197 & 0.202 & 0.474 & 0.356 & 0.884 & 0.370 & 0.977 \\
Rep, Woman & 0.198 & 0.201 & 0.506 & 0.351 & 0.897 & 0.329 & 0.978 \\
Ind, Man & 0.198 & 0.200 & 0.510 & 0.349 & 0.897 & 0.322 & 0.983 \\
Ind, Woman & 0.201 & 0.202 & 0.517 & 0.347 & 0.901 & 0.315 & 0.989 \\
\end{tabular}
\caption{Median permutation scores for in-group and cross-group cohesion on $\mathcal{D}_\textit{voiced}$ after CrowdTruth filtering}
\label{tab:tb_ct_median_metrics}
\end{table*}

\begin{table*}
\centering
\small
\begin{tabular}{
    r|ccccccc
}
& & & & Cross $\otimes$ & Plurality $\cap$ & Voting $\otimes$ & \\
Group & {IRR $\cap$} & {XRR $\otimes$} & {Negentropy $\cap$} & {Negentropy} & {Size} & {Agreement} & {GAI} \\
\hline
Dem & 0.304 & 0.304 & 0.541 & 0.515 & 0.903 & 0.326 & 1.001 \\
Rep & 0.304 & 0.303 & 0.601 & 0.466 & 0.938 & 0.327 & 1.004 \\
Ind & 0.300 & 0.302 & 0.601 & 0.465 & 0.938 & 0.327 & 0.998 \\
\hline
Men & 0.304 & 0.304 & 0.539 & 0.518 & 0.902 & 0.326 & 1.000 \\
Women & 0.303 & 0.304 & 0.511 & 0.544 & 0.886 & 0.326 & 0.998 \\
\hline
Dem, Men & 0.303 & 0.304 & 0.633 & 0.443 & 0.959 & 0.331 & 1.002 \\
Dem, Women & 0.302 & 0.303 & 0.607 & 0.461 & 0.942 & 0.328 & 0.995 \\
Rep, Men & 0.296 & 0.303 & 0.648 & 0.431 & 0.969 & 0.330 & 0.972 \\
Rep, Women & 0.303 & 0.302 & 0.648 & 0.432 & 0.968 & 0.328 & 0.995 \\
Ind, Men & 0.294 & 0.302 & 0.650 & 0.430 & 0.970 & 0.330 & 0.975 \\
Ind, Women & 0.293 & 0.301 & 0.647 & 0.431 & 0.968 & 0.326 & 0.988 \\
\end{tabular}
\caption{Median permutation scores for in-group and cross-group cohesion on $\mathcal{D}_\textit{toxicity}$ after CrowdTruth filtering}
\label{tab:tb_ct_median_metrics_tr}
\end{table*}

\begin{table*}
\centering
\small
\begin{tabular}{
    r|ccccccc
}
& & & & Cross $\otimes$ & Plurality $\cap$ & Voting $\otimes$ & \\
Group & {IRR $\cap$} & {XRR $\otimes$} & {Negentropy $\cap$} & {Negentropy} & {Size} & {Agreement} & {GAI} \\
\hline
Rep $\rightarrow$ Dem (v Dem) & 0.208 & 0.183 & 0.492 & 0.433 & 0.899 & 0.337 & 1.130 \\
Ind $\rightarrow$ Dem (v Dem) & 0.211 & 0.184 & 0.511 & 0.433 & 0.905 & 0.321 & 1.137 \\
\hline
Dem $\rightarrow$ Rep (v Rep) & 0.202 & 0.172 & 0.448 & 0.406 & 0.871 & 0.264 & 1.171 \\
Ind $\rightarrow$ Rep (v Rep) & 0.202 & 0.170 & 0.467 & 0.405 & 0.879 & 0.253 & 1.179 \\
\hline
Dem $\rightarrow$ Ind (v Ind) & 0.171 & 0.186 & 0.444 & 0.430 & 0.872 & 0.319 & 0.921 \\
Rep $\rightarrow$ Ind (v Ind) & 0.170 & 0.186 & 0.439 & 0.431 & 0.871 & 0.328 & 0.915 \\
\end{tabular}
\caption{Median permutation scores for vicarious alignment on $\mathcal{D}_\textit{voiced}$ after CrowdTruth filtering}
\label{tab:tb_ct_vic_median_metrics}
\end{table*}

\end{document}